\documentclass{sigkddExp}
\usepackage{cite}
\usepackage{graphicx}
\usepackage{algorithmic}
\usepackage{multicol}
\usepackage{multirow}
\usepackage{booktabs}
\usepackage{amsmath,amssymb,amsfonts}
\usepackage{microtype}
\usepackage{verbatim}
\usepackage{fancyvrb}
\usepackage{fvextra} 
\usepackage{inconsolata}
\usepackage{subfigure}
\usepackage{hyperref}
\usepackage{enumitem} 
\usepackage{siunitx} 
\begin{document}
%

\title{Time Series Forecasting with LLMs: \\ Understanding and Enhancing  Model Capabilities}
%

%


\author{
%
Hua Tang$^{2,*}$, Chong Zhang$^{3,*}$, Mingyu Jin$^{1}$, Qinkai Yu$^{3}$, Zhenting Wang$^{1}$, \\Xiaobo Jin$^{5}$, Yongfeng Zhang$^{1}$, Mengnan Du$^{4}$. \\
\\
${^1}$Rutgers University, ${^2}$Shanghai Jiaotong University, ${^3}$University of Liverpool, \\
${^4}$New Jersey Institute of Technology, ${^5}$Xi'an Jiaotong-Liverpool University.
\\
\small{Hua Tang* and Chong Zhang* contribute equally to this work}
}

\maketitle
\begin{abstract}
Large language models (LLMs) have been applied in many fields and have developed rapidly in recent years. As a classic machine learning task, time series forecasting has recently been boosted by LLMs. Recent works treat large language models as \emph{zero-shot} time series reasoners without further fine-tuning, which achieves remarkable performance. However, some unexplored research problems exist when applying LLMs for time series forecasting under the zero-shot setting. For instance, the LLMs' preferences for the input time series are less understood. In this paper, by comparing LLMs with traditional time series forecasting models, we observe many interesting properties of LLMs in the context of time series forecasting. First, our study shows that LLMs perform well in predicting time series with clear patterns and trends but face challenges with datasets lacking periodicity. This observation can be explained by the ability of LLMs to recognize the underlying period within datasets, which is supported by our experiments. In addition, the input strategy is investigated, and it is found that incorporating external knowledge and adopting natural language paraphrases substantially improve the predictive performance of LLMs for time series. Our study contributes insight into LLMs' advantages and limitations in time series forecasting under different conditions. The code is at \url{https://github.com/MingyuJ666/Time-Series-Forecasting-with-LLMs. }
\end{abstract}

\section{Introduction}
\vspace{12pt}
Recently, large language models (LLMs) have been widely used and have achieved promising performance across various domains, such as health management, customer analysis, and text feature mining~\cite{peng2023study, ledro2022artificial, huang2023finbert,jin2024impact,jin2024disentangling}. Time series forecasting requires extrapolation from sequential observations. Language models are designed to discern intricate concepts within temporally correlated sequences and intuitively appear well-suited for this task. Hence, some preliminary studies apply LLMs to time series forecasting tasks \cite{gruver2023large, rasul2023lag, sun2023test}. 

However, the application of LLMs for time series forecasting is still in its early stage, and the boundaries of this research area are not yet well defined. There are many unexplored problems in this field. For example, existing research lacks exploration into how the performance of LLMs varies when faced with different types of time series inputs. This includes the effectiveness gap for LLMs in predicting data with seasonal and trending patterns versus data without such patterns. 

To fill this research gap, in this paper, we focus on LLMs' preferences for the input time series in time series forecasting under the zero-shot prompting setting. Experiments on both real and synthesized datasets show that LLMs perform better in time series with higher trend or seasonal strengths. Our observations also reveal that LLMs perform worse when there are multiple periods within datasets, which may be attributed to the fact that LLMs cannot capture distinct periods within those datasets. To further discern the LLMs' preferences for the specific input data segments, we design counterfactual experiments involving systematic permutations of input sequences. The findings suggest that LLMs are particularly sensitive to the segment of input sequences closest to the target output.

Based on the above findings, we want to explore why LLMs forecast well on datasets with higher seasonal strengths. To this end, we require LLMs to tell the period of the datasets through multiple runs. We find that LLMs can mostly recognize the underlying period of a dataset. This can explain the findings of why large language models can forecast time series with high trends or seasonal intensities well since they can obtain the seasonal pattern inside the datasets.

\begin{figure*}
    \centering
    \label{Workflow}
    \vspace{-16pt}
    \includegraphics[width=1.0\linewidth]{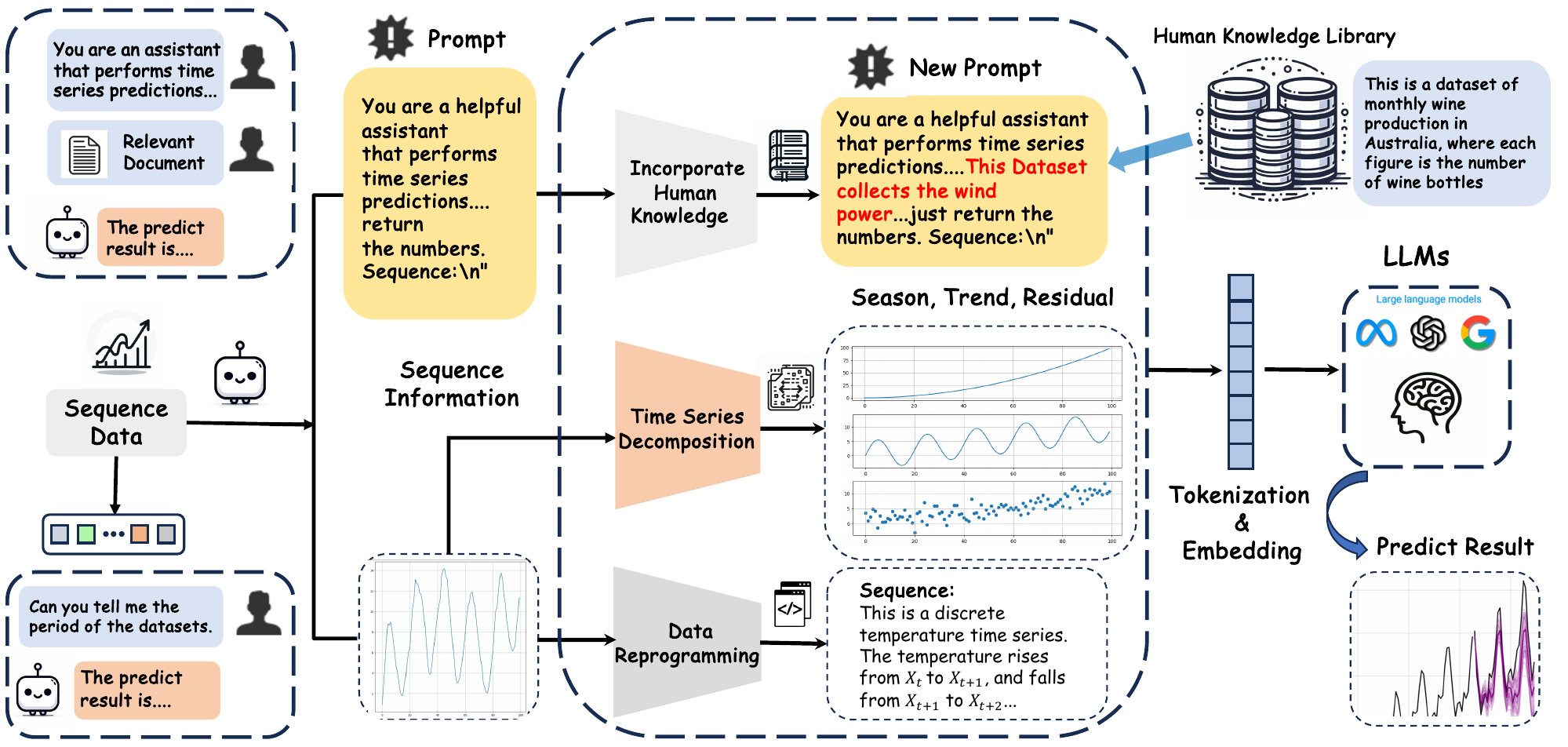}
    \caption{The workflow of our analysis process. Our analysis workflow involves processing sequence data using different tokenization and embedding methods with various LLMs, such as GPTs and Gemini. To analyze the preferences of LLMs, we compute the seasonal and trend strength inside the datasets. Our experiments illuminate that LLMs prefer series with higher seasonal and trend strengths. To elucidate the rationale behind our findings, we demand the LLMs identify the underlying periods, revealing that the model can recognize the underlying periods in most cases. In addition, to improve the performance of time series forecasting, we propose two approaches to the user input: for the input prompt, we incorporate human knowledge regarding the dataset sources, and for the input sequence, we reprogram the data into natural language sequences. Both methods result in substantially improved model performance.}
    \label{fig:workflow}
\end{figure*}

In light of the above-mentioned findings, we are interested in how to leverage these insights to further improve model performance. To address this, we propose two simple techniques to enhance model performance: incorporating external human knowledge and converting numerical sequences into natural language counterparts. Incorporating supplementary information enables large language models to more effectively grasp the periodic nature of time series data, moving beyond a mere emphasis on the tail of the time series. Transforming numerical data into a natural language format enhances the model's ability to comprehend and reason, also serving as a beneficial approach. Both approaches improve model performance and contribute to our understanding of LLMs in time series forecasting. The workflow is illustrated in \autoref{fig:workflow}. 

\noindent The key contributions are as follows:
\begin{itemize}[leftmargin=*]\setlength\itemsep{0.1em}
    \item We investigate the preferences for the input sequences in LLMs in time series forecasting tasks. Our analysis has revealed that LLMs significantly outperform traditional time series forecasting methods without the need for additional fine-tuning. Interestingly, LLMs display superior predictive capabilities when dealing with datasets that have higher trends and seasonal strengths. 
    
    \item We require LLMs to identify the periodicity of datasets across multiple iterations. Our observations indicate that LLMs can effectively recognize the inherent periodic patterns within datasets. This observation answers the question of why LLMs perform well in forecasting time series with higher seasonal strengths, as they can capture the seasonal patterns inherent in the data.
    
    \item We propose two simple techniques to improve model performance and find that both incorporating external human knowledge into input prompts and paraphrasing input sequences to natural language substantially improve the performance of LLMs in time series forecasting.

\end{itemize}

\section{Preliminaries}
\vspace{10pt}
\subsection{Large Language Model}
\vspace{15pt}

We use LLMs as a zero-shot learner for time series forecasting by treating numerical values as text sequences. In this paper, we investigate three close source LLMs, including GPT-3.5-turbo, GPT-4-turbo, and Gemini-1.0-Pro, and one open-source LLMs, i.e., llama-2-13B. The success of LLMs in time series forecasting can significantly depend on correct pre-processing and handling of the data \cite{gruver2023large}. We followed the pre-processing approach of Gruver \cite{gruver2023large} and this process involves the following few steps.

\noindent\textbf{Input Pre-processing.}\,
In this phase of time series forecasting with LLMs, we perform two pre-processing steps. First, numerical values are transformed into strings, a crucial step that significantly influences the model's comprehension and data processing. For instance, a series like 0.123, 1.23, 12.3, 123.0 is reformatted to "1 2, 1 2 3, 1 2 3 0, 1 2 3 0 0", introducing spaces between digits and commas to delineate time steps, while decimal points are omitted to save token space. Second, tokenization is equally important, shaping the model's pattern recognition capabilities. Unlike traditional methods such as byte-pair encoding (BPE) \cite{huggingface_nlp_course_2023}, which can disrupt numerical coherence, we use spacing digits which ensures individual tokenization, enhancing pattern discernment. Third, rescaling is employed to efficiently utilize tokens and manage large inputs by adjusting values so that a specific percentile aligns to 1. This facilitates the model's exposure to varying digit counts and supports the generation of larger values, a testament to the nuanced yet critical nature of data preparation in leveraging LLMs for time series analysis.

\subsection{Time Series Forecasting} 
\label{Time Series Forecasting}
\vspace{15pt}

In the context of time-series forecasting, the primary goal is to predict the values for the next $H$ steps based on observed values from the preceding $K$ steps, which is mathematically expressed as:
\begin{equation}
\hat{X}_{t}, ...,\hat{X}_{t+H-1} = F(X_{t-1}, ..., X_{t-K}; V; \lambda)
\end{equation}
Here, $\hat{X}_{t}, ...,\hat{X}_{t+H-1}$ represent the $H$-step estimation given the previous $K$-step values $X_{t-1}, ..., X_{t-K}$. $\lambda$ denotes the trained parameters from the model $F$, and $V$ denotes the prompt or any other information used for inference. In this paper, we focus predominantly on univariate time series forecasting to investigate the preference and performance of LLMs in univariate \color{black} time series forecasting under the zero-shot setting.

Motivated by interpretability requirements in real-world scenarios, time series can often be decomposed into the trend component, the seasonal component, and the residual component through the addictive model \cite{cleveland1990stl}. The trend component captures the hidden long-term changes in the data, such as the linear or exponential pattern. The seasonal component captures the repeating variation in the data, and the residual component captures the remaining variation in the data after removing the trend and seasonal components. This decomposition offers a method to quantify the properties of time series, which is detailed in \autoref{LLMs' preferences}.

\noindent\textbf{Datasets.} 
In this study, we primarily use Darts \cite{Darts2022Herzen}, a benchmark univariate dataset widely recognized in deep learning research, along with many baseline methods. Darts consists of eight real univariate time series datasets, including those with clear patterns, such as the AirPassengerDataset, and irregular datasets, such as the SunspotsDataset. Besides, we employ some other commonly used datasets, such as US Births Dataset\cite{Monash2021Godahewa}, TSMC-Stock and Turkeypower datasets \cite{gruver2023large} and ETT~\cite{zhou2021informer} in Sections \ref{External Knowledge} and \ref{Naural Language Paraphrasing} to demonstrate the effectiveness of our proposed methods. 
A full description of those datasets can be seen in Section \ref{dataset_prompt}.

\noindent\textbf{Evaluation Metrics.} 
In this paper, we evaluate model performance with three metrics: Mean Squared Error (MSE), Mean Absolute Error (MAE), and Mean Absolute Percentage Error (MAPE). These metrics are defined as follows:
\begin{align}
    &\text{MSE} = \frac{1}{n} \sum_{i=1}^n (y_i - \hat{y}_i)^2\\
    &\text{MAE} = \frac{1}{n} \sum_{i=1}^n \left| y_i - \hat{y}_i \right|\\
    &\text{MAPE} =\frac{1}{n} \sum_{i=1}^n\left|\frac{y_i-\hat{y}_i}{y_i}\right|    
\end{align}
where \( y_i \) denotes the true value, \( \hat{y}_i \) represents the predicted value, and \( n \) is the sample size.

\section{What are LLMs' Preferences in Time Series Forecasting?}
\vspace{15pt}

To explore the preference of LLMs, we first quantify the properties of the input time series to investigate the LLMs' preferences for time series. Then, to further emphasize our findings, we evaluate the importance of different segments of the input sequence by adding Gaussian noise to the original time series. 

\subsection{Analyzing Method}
\vspace{15pt}

We first compare the performance between LLMs and traditional time series forecasting methods, as shown in \autoref{tab:comparison-with-traditional-methods}. It is shown that LLMs perform better within most datasets. 
GPT-4-turbo and Llama-2 perform relatively well on the AirPassengerdataset and the AusBeerdataset with low  MAPE. 
Gemini outperforms GPT-3.5-turbo on time series forecasting and outperforms GPT-4-turbo on some datasets but is on par with GPT-4-turbo overall.

To understand the preferences of the LLMs, we compare our framework using various foundational models, such as GPT-4-turbo and GPT-3.5-turbo, with traditional methods. We also design experiments on synthesized datasets to validate our findings and analyze the impact of the multiple periods. To quantify the LLMs' preferences towards time series, following \cite{wang2006characteristic}, we define the strength of the trend and the seasonality as follows:

\begin{equation}
    \begin{aligned}
        & Q_T \hspace{-0.2em}=\hspace{-0.2em} 1\hspace{-0.2em}-\hspace{-0.2em}\frac{\text{Var}(X_{R})}{\text{Var}(X_{T}\hspace{-0.2em}+\hspace{-0.2em}X_{R})},\hspace{-0.2em}
        & Q_S \hspace{-0.2em}=\hspace{-0.2em} 1\hspace{-0.2em}-\hspace{-0.2em}\frac{\text{Var}(X_{R})}{\text{Var}(X_{S}\hspace{-0.2em}+\hspace{-0.2em}X_{R})}
    \end{aligned}
\end{equation}
where $X_{K}\in {R}^{K}$, $X_{S}\in {R}^{K}$ and $X_{R}\in {R}^{K}$ denote the trend component, the seasonal component and the residual component respectively. The presented indices indicate the trend's strength and seasonality, providing a measure ranging up to 1. It is easy to find that a higher value indicates a stronger trend or seasonality within the time series. 
Throughout this paper, we use the word "higher strength" to represent the comparison of the strengths between different datasets. The assessment of strength is not based on a fixed level, as the concepts of "strong" and "weak" vary across different datasets and scenarios.

To further discern the LLMs' preferences for the specific segments of the input data, we add Gaussian noise to the original time series to create counterfactual examples. We start by defining a sliding window that constitutes 10\% of the total length of the time series, and we set the sliding window to gradually move closer to the output sequence. This method allows us to assess the impact of different segments fairly and thereby infer the interpretability of the time series segments that LLMs predominantly focus on. 

\subsection{Preferences for Input Sequences}
\label{LLMs' preferences}
\vspace{15pt}


In this subsection, we investigate the input sequence preferences for time series forecasting with LLMs. 
We conduct experiments on real datasets with GPT-3.5-turbo and GPT-4-turbo, measuring model performance through MAPE. To further validate our findings, we also use GPT-3.5-turbo and Gemini-1.0-Pro to forecast multiple-period time series on synthesized datasets.

\begin{table*}[!ht]
   \centering
   \scriptsize
   \vspace{-10pt}
   \caption{Correlation matrix between the strengths of the input time series and the model performance.}
   \begin{tabular}{l|S S|S S}
    \hline
    \textbf{Metrics} & \textbf{GPT4-MAPE}  & \textbf{GPT3.5-MAPE} & \textbf{Trend Strength $Q_T$} & \textbf{Seasonal Strength $Q_S$} \\
    \hline
    \textbf{GPT4-MAPE}  & 1.00000 & 0.987398  & -0.020637 & -0.681440 \\
    \textbf{GPT3.5-MAPE}  & 0.987398 & 1.00000 & -0.115087 & -0.669983 \\
    \textbf{Trend Strength $Q_T$} & -0.020637 & -0.115087  & 1.00000 & 0.508980 \\
    \textbf{Seasonal Strength $Q_S$} & -0.681440  & -0.669983 & 0.508980 & 1.00000 \\
    \hline
  \end{tabular}
  \label{Pearson_correlation_matrix_input}
\end{table*}

\subsubsection{Implementation Details}
\vspace{15pt}
\noindent\textbf{Real Datasets:}\,
We conduct experiments on ten real-world datasets, including both those with clear patterns and those with irregular characteristics. The results are shown in \autoref{tab: Q1}.
We apply the Seasonal-Trend decomposition using the LOESS (STL) technique \cite{cleveland1990stl} to decompose the original time series into trend, seasonal, and residual components. Subsequently, we compute the strengths of the trend strength $Q_T$ and seasonal strength $Q_S$. To further understand the LLMs' preferences for the specific segments of the input data, we conduct the counterfactual analysis with a systematic permutation to the input time series. We first scale the sequence through max-min normalization. We then define a sliding window that constitutes 10\% of the total length of the time series and add Gaussian noise into the data within this window data. Subsequently, the sliding window moves closer to the last known data point.

\noindent\textbf{Synthesized Datasets:}\,
To further validate our findings and investigate the influence of the number of periods on model performance, we generate a dataset using the function $y = \alpha *x + \beta_1*cos (2\pi f_1*x) + \beta_2*cos (2\pi f_2*x)+ \epsilon$. $x$ ranges from 0 to 20 and $\epsilon$ follows the normal distribution $\mathcal{N}(0, 1)$. 

\subsubsection{Key Findings}
\begin{figure*}[!ht]
    \subfigure[TSMCStock]{\includegraphics[width=0.33\linewidth]{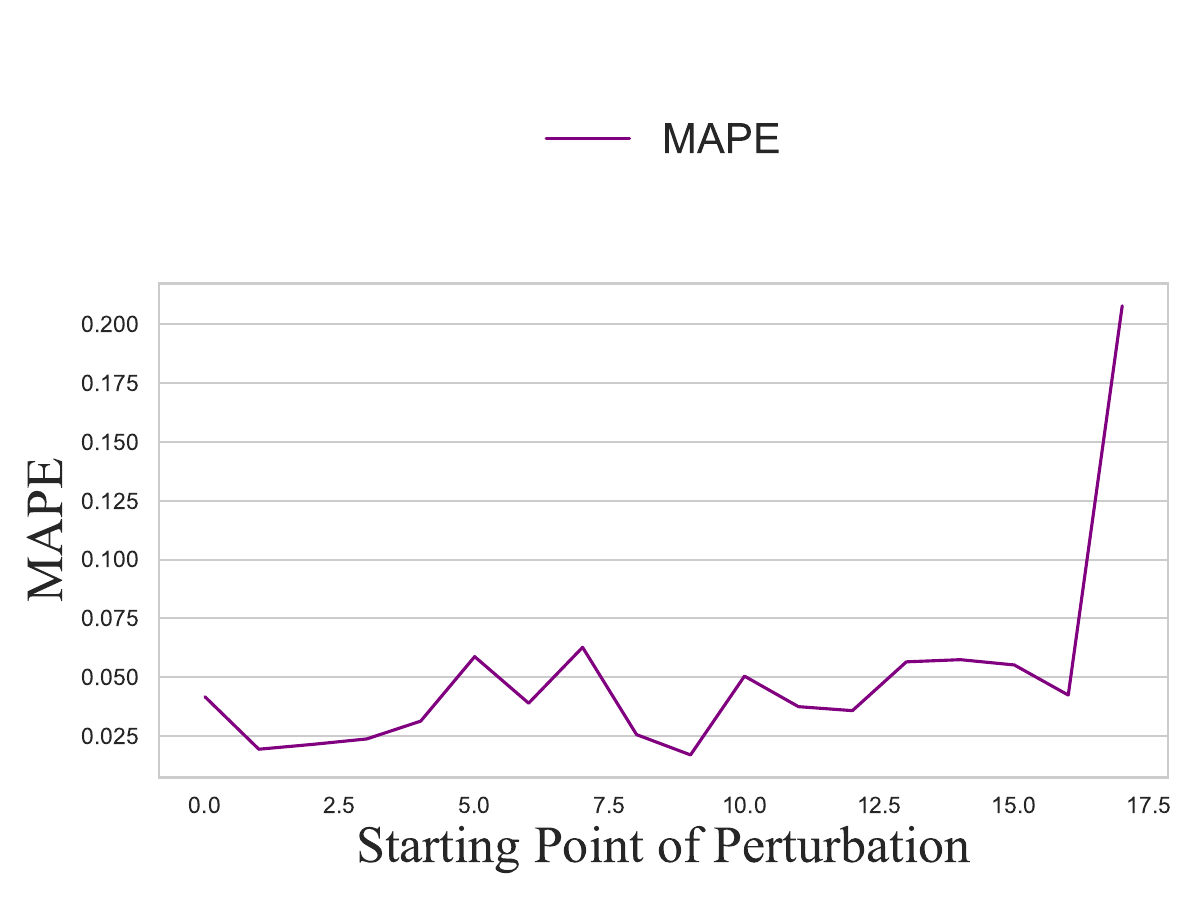}}
    \subfigure[IstanbulTrafficDataset]{\includegraphics[width=0.33\linewidth]{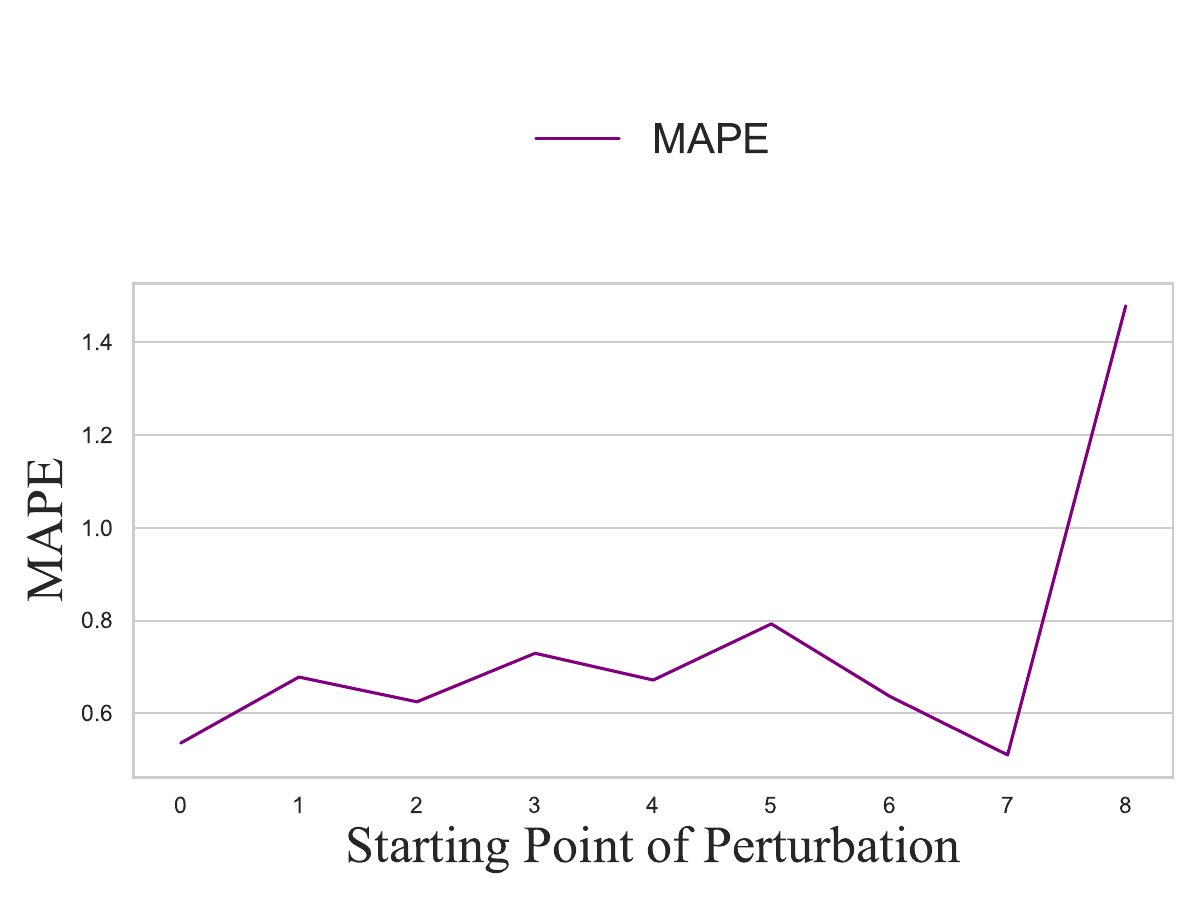}}
    \subfigure[MonthlyMilk]{\includegraphics[width=0.33\linewidth]{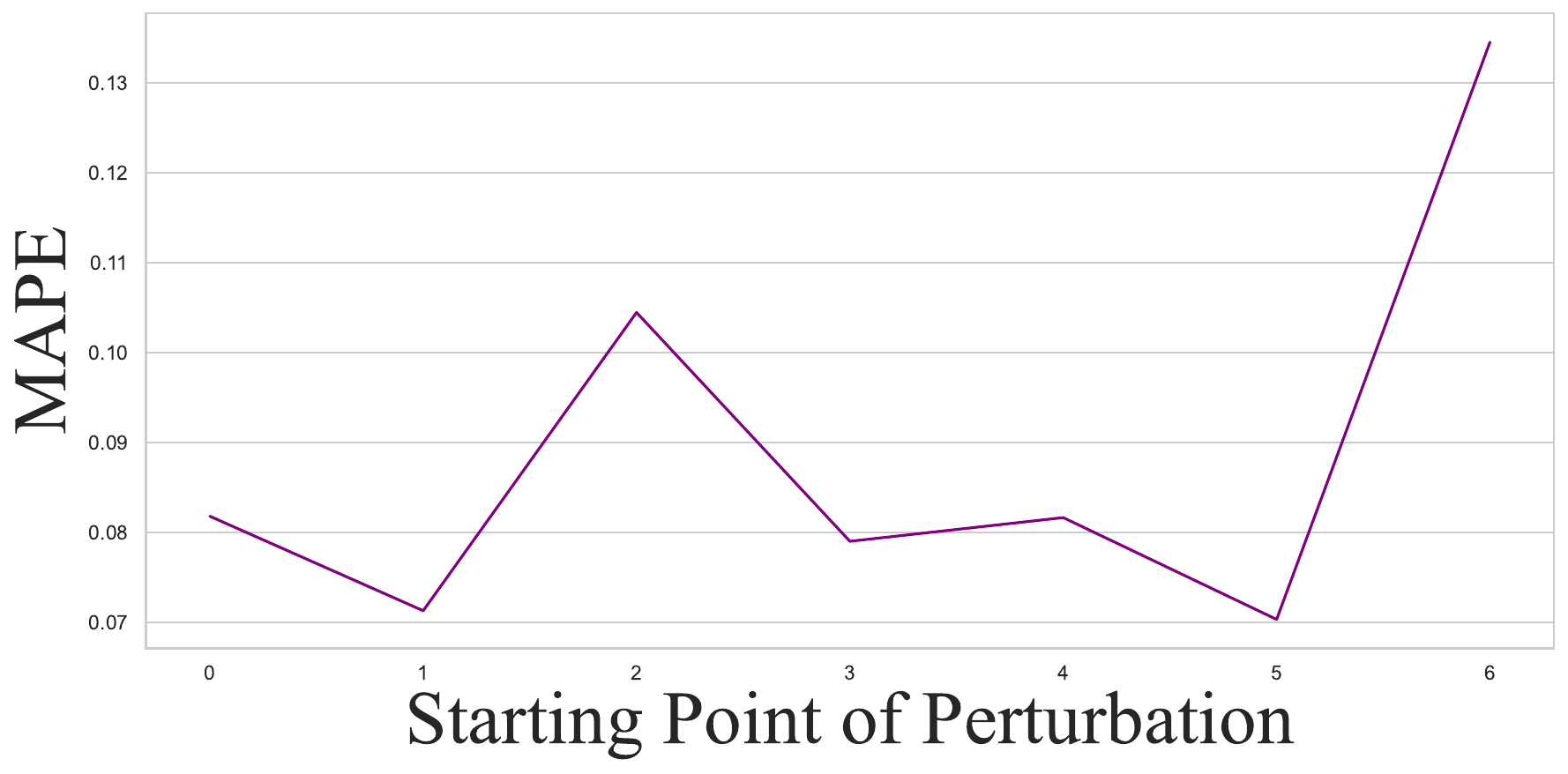}}
    \caption{Experiments of Sequence Focused Attention Through Counterfactual Explanation on GPT-3.5-turbo. 
    }
    \label{Counterfactual analysis on GPTs}
\end{figure*}

\begin{figure*}[!ht]
    \subfigure[TSMCStock]{\includegraphics[width=0.33\linewidth]{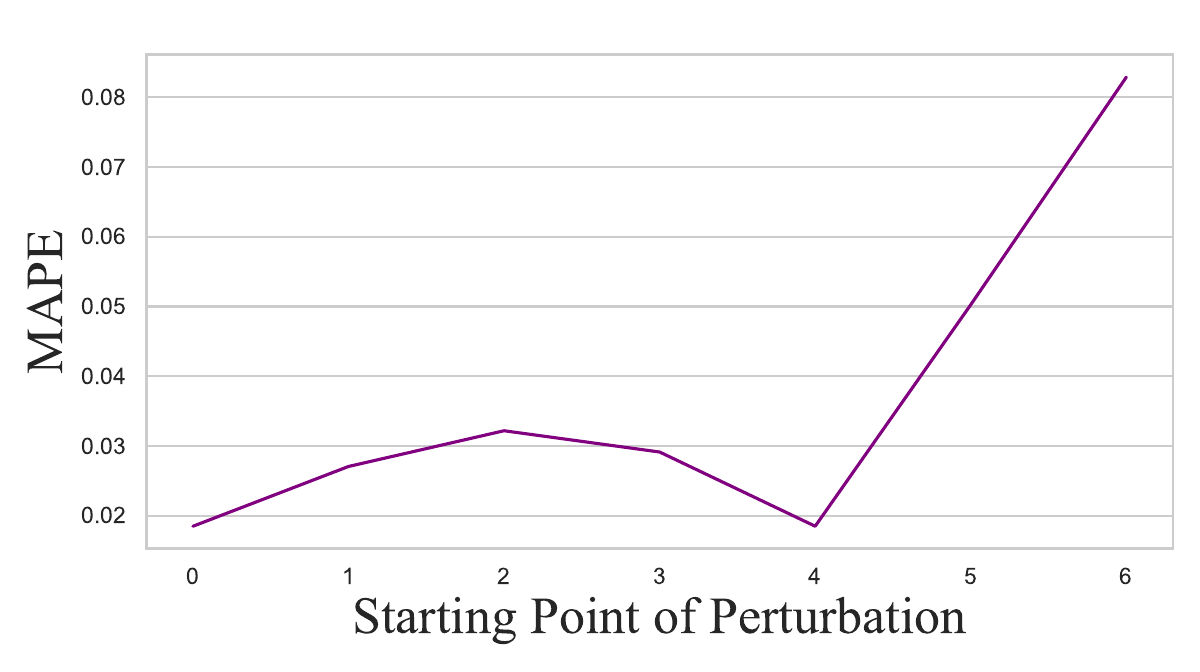}}
    \subfigure[IstanbulTrafficDataset]{\includegraphics[width=0.33\linewidth]{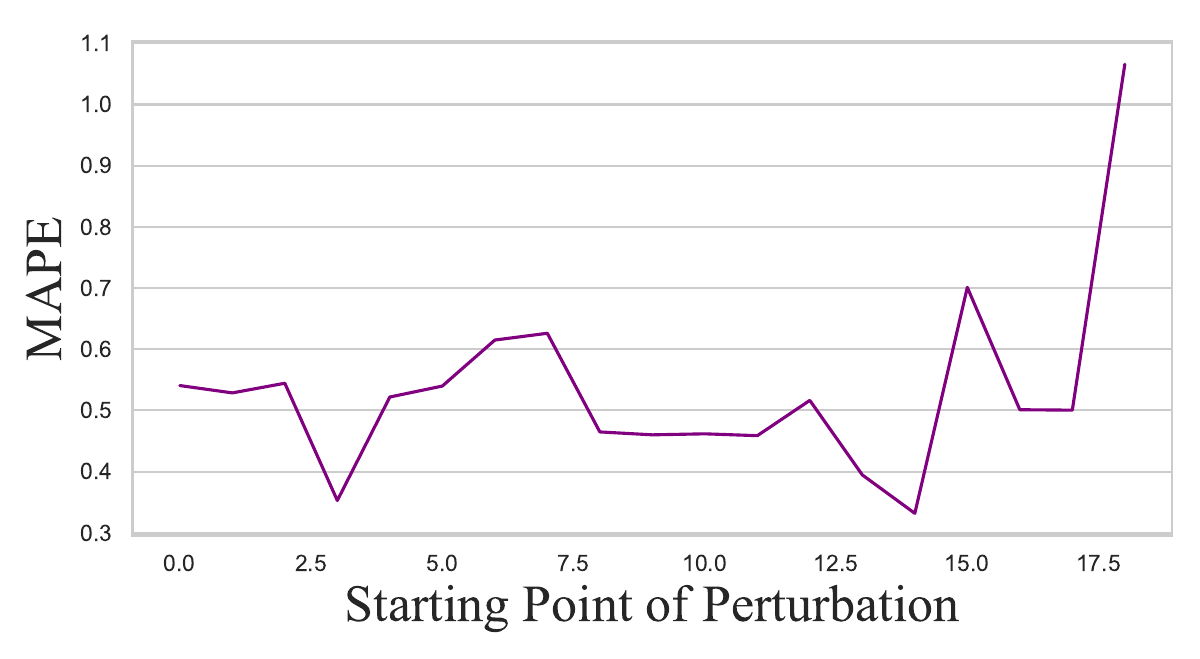}}
    \subfigure[MonthlyMilk]{\includegraphics[width=0.33\linewidth]{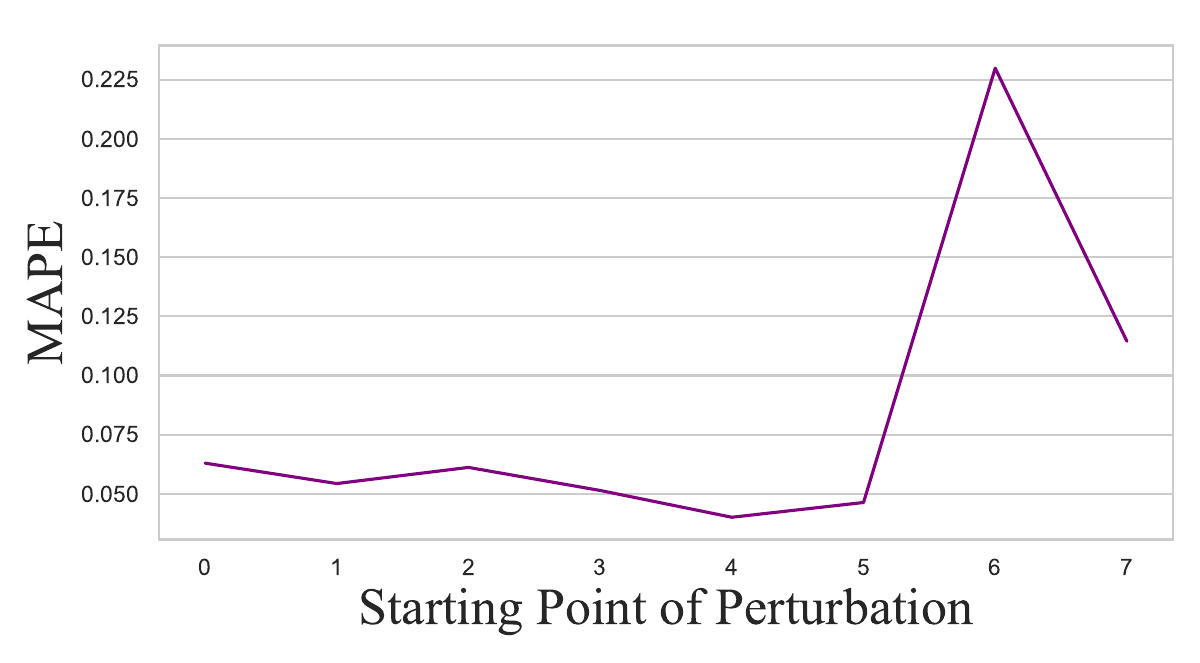}}
    \caption{Experiments of Sequence Focused Attention Through Counterfactual Explanation on Gemini-Pro-1.0.}
    \label{Counterfactual analysis on Gemini}
\end{figure*}

\vspace{15pt}
After computing the Pearson correlation coefficients (PCC), we observe a nearly strong correlation between the strengths and model performance, showing that LLMs perform better when the input time series has a higher trend and seasonal strength, which is shown in \autoref{Pearson_correlation_matrix_input}.
In the context of multi-period time series, the model performance worsens as the number of periods increases. It indicates that LLMs may have difficulty recognizing the multiple periods inherent in such datasets. Besides, for counterfactual analysis, as shown in \autoref{Counterfactual analysis on GPTs} and \autoref{Counterfactual analysis on Gemini}, there is a noticeable increase in MAPE values when Gaussian noise is added to the latter segments, 
while the perturbation of the first part of the sequence has little effect on the prediction performance. Our findings reveal that LLMs are more sensitive to the end of input time series when forecasting. We show our full results in \autoref{Counterfactual analysis on GPTs} and \autoref{Counterfactual analysis on Gemini}. As we move to the right along the x-axis, the closer it gets to the output sequence. 
It is also found that the initial part of the sequence has the least impact on the prediction accuracy. For the datasets with high seasonal strengths over 85\%, such as WoolyDataset, and MonthlymilkDataset, more than 80\% of the length of the time series has almost no effect on the model performance. 

\section{Why do LLMs Forecast Well on Data with Higher \\ Seasonal Strengths?}
\vspace{15pt}
Our findings show that LLMs demonstrate enhanced performance in time series forecasting with strong seasonal strengths. This raises the question: Why do LLMs perform well in forecasting datasets with marked seasonal patterns? To explore this phenomenon, we craft prompts that require LLMs to recognize the dataset's temporal pattern.

This approach is grounded in the hypothesis that LLMs are proficient in handling datasets with distinct seasonal attributes. By explicitly prompting LLMs to predict the dataset's period, we aim to leverage their inherent ability to discern and extrapolate from complex patterns, which sheds light on the mechanisms that underpin their superior performance in such contexts.

\begin{table*}[htbp]
    \caption{Comparison test of traditional prediction methods.}
    \centering
    \scriptsize
    \begin{tabular}{|l|l||l|l|}
        \hline
        \multicolumn{2}{|c||}{\textbf{AirPassengers}} & \multicolumn{2}{c|}{\textbf{AusBeer}} \\
        \hline
        \textbf{Method} & \textbf{MSE / MAE / MAPE} & \textbf{Method} & \textbf{MSE / MAE / MAPE} \\
        \hline
        Exponential Smoothing & 2007.67 / 37.91 / 8.10 & Exponential Smoothing & 703.26 / 22.80 / 5.44 \\
        SARIMA & 2320.47 / 39.80 / 8.46 & SARIMA & \textbf{475.53} / 19.07 / 4.49 \\
        Cyclical Regression & 2028.37 / 36.70 / 8.52 & Cyclical Regression & 989.31 / 26.29 / 6.13 \\
        AutoARIMA & 8702.09 / 68.52 / 13.98 & AutoARIMA & 550.05 / 18.84 / 4.41 \\
        FFT & 3274.46 / 46.38 / 10.59 & FFT & 7682.56 / 73.74 / 17.44 \\
        StatsForecastAutoARIMA & 2952.52 / 45.41 / 9.71 & StatsForecastAutoARIMA & 559.46 / 20.56 / 4.86 \\
        Naive Mean & 47703.65 / 204.25 / 44.61 & Naive Mean & 1885.72 / 30.66 / 6.68 \\
        Naive Seasonal & 6032.80 / 62.87 / 14.18 & Naive Seasonal & 10828.02 / 96.35 / 23.39 \\
        Naive Drift & 6505.79 / 72.21 / 17.50 & Naive Drift & 18507.61 / 128.23 / 30.91 \\
        Naive Moving Average & 6032.80 / 62.87 / 14.18 & Naive Moving Average & 10828.02 / 96.35 / 23.39 \\
        N-Beats & 3994.55 / 54.95 / 12.81 & N-Beats & 250.61 / 14.42 / 3.53 \\
        DeepAR & 184222.64 / 421.99 / 98.42 & DeepAR & 16197.17 / 40.23 / 9.89 \\
        Prophet & 7345.31 / 43.87 / 8.62 & Prophet & 6323.89 / 28.76 / 6.92 \\
        LLMTime with GPT-3.5-Turbo & 6244.07 / 61.39 / 14.43 & LLMTime with GPT-3.5-Turbo & 841.68 / 23.59 / 5.62 \\
        LLMTime with GPT-4-Turbo & 1317.9 / 55.49 / 11.18 & LLMTime with GPT-4-Turbo & 513.49 / 18.57 / 4.28 \\
        LLMTime with Gemini-1.0-pro & 6392.21 / 63.57 / 14.03 & LLMTime with Gemini-1.0-pro & 397.78 / 14.36 / 3.27 \\
        LLMtime with Llama-2 & 1286.25 / 28.04 / 6.07 & LLMtime with Llama-2 & 644.82 / 17.88 / 4.08 \\
        \hline
        \hline
        \multicolumn{2}{|c||}{\textbf{MonthlyMilk}} & \multicolumn{2}{c|}{\textbf{Sunspots}} \\
        \hline
        \textbf{Method} & \textbf{MSE / MAE / MAPE} & \textbf{Method} & \textbf{MSE / MAE / MAPE} \\
        \hline
        Exponential Smoothing & 564.94 / 20.23 / 2.41 & Exponential Smoothing & 326750.49 / 499.78 / 3129.63 \\
        SARIMA & 1289.76 / 32.78 / 3.87 & SARIMA & 2902.72 / 45.75 / 466.99 \\
        Cyclical Regression & 3631.53 / 56.15 / 6.60 & Cyclical Regression & 3917.76 / 47.84 / 274.31 \\
        AutoARIMA & 2682.67 / 42.82 / 5.20 & AutoARIMA & 4695.67 / 58.47 / 709.23 \\
        FFT & 3453.96 / 45.62 / 5.48 & FFT & 3784.56 / 49.81 / 150.32 \\
        StatsForecastAutoARIMA & \textbf{186.14} / 10.64 / 1.28 & StatsForecastAutoARIMA & 8406.55 / 72.99 / 95.18 \\
        Naive Mean & 19893.07 / 127.33 / 14.46 & Naive Mean & 4120.40 / 49.84 / 267.22 \\
        Naive Seasonal & 4870.40 / 56.00 / 6.31 & Naive Seasonal & 4440.63 / 56.78 / 688.58 \\
        Naive Drift & 3998.11 / 56.06 / 6.52 & Naive Drift & 5032.77 / 60.40 / 724.88 \\
        Naive Moving Average & 4870.40 / 56.00 / 6.31 & Naive Moving Average & 4440.63 / 56.78 / 688.58 \\
        N-Beats & 3140.89 / 51.57 / 6.07 & N-Beats & 4877.59 / 56.58 / 105.55 \\
        DeepAR & 728289.50 / 851.30 / 99.22 & DeepAR & 3421.02 / 48.93 / 132.76 \\
        Prophet & 663.41 / 25.76 / 2.92 & Prophet & 6303.57 / 76.83 / 67.97 \\
        LLMTime with GPT-3.5-Turbo & 7507.13 / 66.28 / 112.77 & LLMTime with GPT-3.5-Turbo & 6556.55 / 58.95 / 217.94 \\
        LLMTime with GPT-4-Turbo & 4442.18 / 50.75 / 172.82 & LLMTime with GPT-4-Turbo & 3374.70 / 41.87 / 321.11 \\
        LLMTime with Gemini-1.0-pro & 628.98 / 17.01 / 1.99 & LLMTime with Gemini-1.0-pro & 626.03 / 14.94 / 1.73 \\
        LLMtime with Llama-2 & 3410.20 / 41.40 / 240.25 & LLMtime with Llama-2 & 4467.67 / 48.95 / 91.79 \\
        \hline
        \hline
        \multicolumn{2}{|c||}{\textbf{WineDataset}} & \multicolumn{2}{c|}{\textbf{WoolyDataset}} \\
        \hline
        \textbf{Method} & \textbf{MSE / MAE / MAPE} & \textbf{Method} & \textbf{MSE / MAE / MAPE} \\
        \hline
        Exponential Smoothing & 23709576.52 / 3370.78 / 14.23 & Exponential Smoothing & 24925885.81 / 3548.19 / 14.98 \\
        SARIMA & 1150166.94 / 966.57 / 20.76 & SARIMA & 812352.21 / 759.07 / 16.37 \\
        Cyclical Regression & 7873785.27 / 2148.24 / 8.52 & Cyclical Regression & 1032574.82 / 962.72 / 22.14 \\
        AutoARIMA & 698661.90 / 646.03 / 14.07 & AutoARIMA & 838852.91 / 786.25 / 16.84 \\
        FFT & 1031170.45 / 867.83 / 18.60 & FFT & 1012255.35 / 945.20 / 20.80 \\
        StatsForecastAutoARIMA & 20040877.37 / 2853.17 / 12.05 & StatsForecastAutoARIMA & 917617.19 / 858.57 / 18.91 \\
        Naive Mean & 11557786.19 / 2200.04 / 8.80 & Naive Mean & 816762.31 / 764.73 / 16.12 \\
        Naive Seasonal & 879447.22 / 724.23 / 15.52 & Naive Seasonal & 1051110.81 / 982.25 / 22.19 \\
        Naive Drift & 9609576.04 / 1833.38 / 7.36 & Naive Drift & 812352.21 / 759.07 / 16.37 \\
        Naive Moving Average & 9070696.99 / 1719.17 / 6.90 & Naive Moving Average & 1032574.82 / 962.72 / 22.14 \\
        N-Beats & 5418377.00 / 1887.30 / 7.68 & N-Beats & 653104.31 / 743.54 / 15.96 \\
        DeepAR & 715027008.00 / 26236.14 / 89.91 & DeepAR & 243831.14 / 4897.85 / 94.89 \\
        Prophet & 4846922.27 / 2201.57 / 8.27 & Prophet & 365241.98 / 891.70 / 34.65 \\
        LLMTime with GPT-3.5-Turbo & 30488.60 / 388.28 / 15.83 & LLMTime with GPT-3.5-Turbo & 526903.08 / 574.58 / 12.00 \\
        LLMTime with GPT-4-Turbo & 22488.17 / 253.08 / 9.98 & LLMTime with GPT-4-Turbo & 942987.19 / 871.64 / 18.55 \\
        LLMTime with Gemini-1.0-pro & 258584.78 / 3645.23 / 14.60 & LLMTime with Gemini-1.0-pro & 64.92 / 6.39 / 7.04 \\
        LLMtime with Llama-2 & 951194.94 / 240.08 / 9.45 & LLMtime with Llama-2 & 675062.52 / 736.04 / 15.83 \\
        \hline
        \hline
        \multicolumn{2}{|c||}{\textbf{HeartRateDataset}} & \multicolumn{2}{c|}{\textbf{Weather}} \\
        \hline
        \textbf{Method} & \textbf{MSE / MAE / MAPE} & \textbf{Method} & \textbf{MSE / MAE / MAPE} \\
        \hline
        Exponential Smoothing & 11.16 / 1.38 / 1.49 & Exponential Smoothing & 1684.38 / 31.60 / 6.79 \\
        SARIMA & 12.98 / 1.34 / 1.61 & SARIMA & 1943.81 / 33.33 / 7.09 \\
        Cyclical Regression & 13.58 / 1.31 / 1.20 & Cyclical Regression & 1700.73 / 30.77 / 7.15 \\
        AutoARIMA & 13.26 / 1.25 / 1.39 & AutoARIMA & 7315.10 / 57.44 / 11.70 \\
        FFT & 13.95 / 1.16 / 1.34 & FFT & 2752.02 / 38.90 / 8.87 \\
        StatsForecastAutoARIMA & 10.53 / 1.27 / 1.39 & StatsForecastAutoARIMA & 2479.55 / 38.06 / 8.16 \\
        Naive Mean & 12.02 / 1.27 / 1.26 & Naive Mean & 39879.84 / 168.27 / 36.44 \\
        Naive Seasonal & 10.55 / 1.32 / 1.31 & Naive Seasonal & 5057.47 / 52.81 / 11.89 \\
        Naive Drift & 10.60 / 1.15 / 1.30 & Naive Drift & 5466.23 / 60.58 / 14.70 \\
        Naive Moving Average & 12.13 / 1.27 / 1.34 & Naive Moving Average & 5057.47 / 52.81 / 11.89 \\
        N-Beats & 72.11 / 7.10 / 7.40 & N-Beats & 4532.84 / 39.21 / 23.49 \\
        DeepAR & 286.82 / 15.67 / 16.36 & DeepAR & 6325.75 / 35.97 / 16.59 \\
        Prophet & 88.93 / 10.97 / 6.54 & Prophet & 3768.15 / 29.36 / 24.01 \\
        LLMTime with GPT-3.5-Turbo & 76.83 / 7.15 / 7.42 & LLMTime with GPT-3.5-Turbo & 224.54 / 3.07 / 0.83 \\
        LLMTime with GPT-4-Turbo & 988.14 / 26.57 / 29.22 & LLMTime with GPT-4-Turbo & 111.65 / 2.40 / 0.64 \\
        LLMTime with Gemini-1.0-pro & 57.96 / 6.01 / 6.66 & LLMTime with Gemini-1.0-pro & 176.32 / 3.72 / 0.75 \\
        LLMtime with Llama-2 & 75.58 / 7.11 / 7.94 & LLMtime with Llama-2 & 215.39 / 4.07 / 1.31 \\
        \hline
    \end{tabular}
    \label{tab:comparison-with-traditional-methods}
\end{table*}

\subsection{Implementation Details}
\vspace{15pt}
To explore the phenomenon that LLMs forecast well on datasets with higher seasonal strengths, we design experiments to verify this phenomenon. We tokenize the input sequence and let the LLMs output the period directly. 
We use GPT-3.5-turbo, GPT-4-turbo, and Gemini-1.0-Pro to predict the periods.
We have chosen five datasets 
with their seasonal strengths exceeding 85\%. 
These datasets are readily available with clear seasonal patterns. In contrast, determining the specific periods of other irregular datasets is challenging, as they have no specific cycles. 
We record the predicted periods ten times and identify the mode period, which is the most frequently predicted value. 
We then compare the mode of these ten results with the real period. The mode is selected as the evaluation metric because, when considering the usage characteristics of LLMs, the output of this number best represents the model's normal performance. The results are shown in \autoref{tab:tab_period}.

\subsection{Key Findings}
\vspace{15pt}
According to the results, we find that large language models can mostly determine the periodicity of a dataset. 
The true periods are determined here by the periodogram, which is commonly used to identify the dominant periods \cite{bartlett1950periodogram}.
The multiples of the predicted period also align with the original data cycle. Consequently, we consider the prediction of these multiples to be accurate. We observe that LLMs generally perform well in predicting the period for most datasets with minimal fluctuations. Surprisingly, we discover that in the case of WoolyDataset and AusbeerDataset, which possess relatively short underlying periods, the predicted period is consistently 3 instead of the true period, 4. This discrepancy may be attributed to the LLMs' tendency to focus on cyclic patterns among individual digits rather than considering the entire sequence as a whole, a phenomenon that could also be interpreted as the model's identification of the underlying cycle. We leave a comprehensive analysis of this phenomenon in the future. 

\vspace{-5pt}

\section{How to Leverage These Insights to Improve the Model's Performance?}
\vspace{15pt}
Based on the findings in the previous two sections, our focus is now on how to leverage these findings to further improve model performance. In this paper, we propose two approaches to the user input without additional fine-tuning: for the input prompt, we incorporate additional knowledge of the specific trend and seasonal patterns in the dataset, which gives the model a richer understanding of the underlying patterns. Regarding the input sequence, we transform the time series data into formats resembling natural language sequences rather than relying on the original tokenization. This approach leverages LLMs' superior capabilities with language sequences. Both methods achieve substantially improved model performance. 

\subsection{Dataset description and the External Knowledge incorporated in the Prompts}
\label{dataset_prompt}
\vspace{15pt}

We briefly introduce the datasets we use, which also serve as the external knowledge incorporated into the prompts. Following \cite{gruver2023large}, we downsample the input series to an hourly frequency, yielding a total of 267 observations and resulting in relatively small datasets. Additionally, we incorporate Memorization datasets published after September 2021, the cutoff date for GPT-3.5-turbo, to demonstrate the effectiveness of TimeLLM and our proposed methods. Finally, we implemented univariate time series forecasting to predict the 'OT' feature on the ETTh1 and ETTm2 datasets, focusing on the last 96 steps of the test set.

\subsection{External Knowledge Enhancing Time Series Forecasting}
\label{External Knowledge}
\vspace{15pt}

We introduce a novel method to improve the performance of large language models for time series forecasting. The core idea of this part is to use the knowledge obtained from the pre-training stage to help predict. 
We provide the large language model with some basic information about the current dataset, such as the background of the data collection, and this process does not involve data leakage. We incorporate our tests on the data leakage in Section \ref{App: Data leakage test}. 
We do not provide the LLMs with any statistical information, such as the periods or trends. This approach ensures that the LLMs forecast the time series entirely based on the data and their prior knowledge. Let \( V_s \) denote the initial prompt representing the original time sequence, and let \( z \) denote the additional information. Consequently, the new prompt \( V_e \) can be expressed as:
$V_e = z + V_s$.

\subsubsection{Implementation Details}
\vspace{15pt}
We input the dataset's external knowledge through prompts before the sequence's input. The external knowledge of each dataset is presented in \autoref{dataset_prompt}. The results are shown in \autoref{tab: Q1}, 
where LLMTime Prediction refers to the approach described by \cite{gruver2023large} without any modifications.

\subsubsection{Key Findings}
\vspace{15pt}
As shown in \autoref{tab_knowledge_enhance}, this method achieves improved performance in most scenarios. Besides, GPT-4-turbo generally performs better than GPT-3.5-turbo on MSE, MAE, and MAPE, especially on AirPassengers, AusBeer, and other datasets. Llama-2 significantly outperforms GPT-3.5-turbo and GPT-4-turbo in terms of MSE and MAE metrics on some datasets (e.g., Wooly, ETTh1, ETTm2), indicating that it can capture data features more accurately. 
Using External Knowledge Enhancing, Gemini outperforms other models on MonthlyMilk, Sunspots, Wooly, and HeartRate Datasets, but performs poorly on other datasets.

\subsection{Natural Language Paraphrasing Time Series Forecasting}
\label{Naural Language Paraphrasing}
\vspace{15pt}

In this subsection, we conduct experiments on the natural language paraphrasing of the input time sequences. This strategy capitalizes on the advanced abilities of large language models in handling language sequences. It is motivated by the fact that LLMs are insensitive by the order of magnitude and size of digits \cite{shah-etal-2023-numeric}.

We use natural language to describe the trend between consecutive values. For instance, given a time series $X$ where $X = [X_{1}, X_{2}, X_{3}, \ldots, X_{n}]$, we describe the trend from $X_t$ to $X_{t+1}$ as follows: "The value rises from $X_t$ to $X_{t+1}$, and falls from $X_{t+1}$ to $X_{t+2}$...". The string we get here is our natural language paraphrasing sequence. After generating responses based on the string, we extract the values from the text and construct the predicted time series.

\begin{table*}[!ht]
\vspace{-12pt}
\centering
\small
\caption{Results and comparison of time series period prediction based on GPT-3.5-turbo and Gemini.}
\begin{tabular}{c|c|cccccccccc|c|c}
\hline
\textbf{Model} & \textbf{Dataset} & \multicolumn{10}{c|}{\textbf{Period }} & \textbf{Real } & \textbf{Mode}\\ \hline
& \textbf{AirPassengersDataset} & 24 & 24 & 7 & 24 & 12 & 24 & 11 & 24 & 24 & 24 & 12 & 24 \\
& \textbf{WineDataset} & 11 & 12 & 24 & 24 & 24 & 20 & 24 & 24 & 24 & 24 & 12 & 24 \\
\textbf{GPT-3.5-turbo} & \textbf{MonthlyMilkDataset} & 6 & 9 & 12 & 9 & 12 & 12 & 12 & 12 & 12 & 11 & 12 & 12 \\
& \textbf{WoolyDataset} & 4 & 3 & 4 & 3 & 3 & 4 & 3 & 3 & 6 & 3 & 4 & 3 \\ 
& \textbf{AusBeerDataset} & 3 & 3 & 3 & 3 & 3 & 3 & 3 & 3 & 3 & 3 & 4 & 3 \\ \hline

& \textbf{AirPassengersDataset} & 11 & 12 & 12 & 4 & 12 & 12 & 12 & 12 & 12 & 12 & 12 & 12\\
& \textbf{WineDataset} & 10 & 12 & 24 & 12 & 6 & 12 & 12 & 24 & 12 & 12 & 12 & 12 \\
\textbf{Gemini-Pro-1.0} & \textbf{MonthlyMilkDataset} & 16 & 12 & 12 & 12 & 12 & 39 & 12 & 11 & 12 & 12 & 12 & 12\\
& \textbf{WoolyDataset} & 5 & 7 & 4 & 5 & 4 & 4 & 4 & 4 & 5 & 6 & 4 & 4 \\ 
& \textbf{AusBeerDataset} & 4 & 4 & 4 & 2 & 5 & 5 & 4 & 3 & 5 & 7 & 4 & 4 \\ \hline

& \textbf{AirPassengersDataset} & 12 & 12 & 12 & 12 & 12 & 12 & 12 & 12 & 12 & 12 & 12 & 12 \\
& \textbf{WineDataset} & 7 & 6 & 6 & 6 & 7 & 7 & 6 & 6 & 6 & 6 & 12 & 6 \\
\textbf{GPT-4-turbo} & \textbf{MonthlyMilkDataset} & 10 & 12 & 12 & 12 & 12 & 14 & 12 & 12 & 12 & 12 & 12 & 12 \\
& \textbf{WoolyDataset}& 5 & 5 & 7 & 5 & 5 & 5 & 7 & 5 & 5 & 4 & 4 & 5 \\
& \textbf{AusBeerDataset} & 4 & 4 & 4 & 4 & 6 & 4 & 4 & 4 & 6 & 4 & 4 & 4 \\\hline

\end{tabular}
\label{tab:tab_period}
\end{table*}

\subsubsection{Implementation Details}
\vspace{15pt}
We use GPT-3.5-Turbo, GPT-4-turbo, Llama-2 and Gemini-Pro-1.0 to forecast the time series, where part of the results are presented in \autoref{tab_paraseq_res} due to the page limit.

\subsubsection{Key Findings}
\vspace{15pt}
According to the results in \autoref{tab_paraseq_res}, we find that enhancing LLM through natural language paraphrasing improves time series forecasting on most datasets. For instance, GPT-3.5-turbo and GPT-4-turbo perform better on most datasets, especially on Natural Language Paraphrasing methods. Gemini outperforms other LLMs on Wooly and AusBeer datasets but underperforms on others with natural language paraphrasing. All these results demonstrate the superior performance of our methods.

\subsection{Computational Cost}
\vspace{15pt}
For reference, we list the average token length cost associated with external knowledge enhancement and natural language paraphrasing. Avg Token Length(ori) is the prompt Length of the unexecuted method, and Avg Token Length(EKE, NLP) is the prompt length after executing the corresponding policy. It is noted that Natural Language Paraphrasing is judged one by one through hard coding. Besides, there is a length check after transformation, so it is guaranteed that a certain length can be obtained each time. The results are shown in \autoref{table:token_length_comparison}. Several key observations can be made: the original TimeLLM method maintains a uniform token length of 200 across all datasets, providing a stable baseline. EKE results in a slight increase in token length, ranging from 7\% to 12\%, suggesting a good balance between incorporating additional context and maintaining computational efficiency. In contrast, NLP leads to a more substantial increase in token numbers.

\subsection{Tests on Data Leakage}
\label{App: Data leakage test}
\vspace{15pt}

We indirectly explore the data leakage problem by asking LLMs if they can identify the dataset name, the first 20 steps of the predicted dataset, and identify the dataset based on the first 20 steps of the time series data points. The results show that although GPT and Gemini can identify and determine data sets with limited information, they generally do not have detailed sequence data knowledge for a wider range of data sets \autoref{table:dataset_tests}.

\begin{table*}[!ht]
    \small
    \centering
    \caption{The results of natural language paraphrasing of sequences and baseline comparison(Partial).}
    \begin{tabular}{c|c|ccc|ccc}
    \hline
    \multirow{2}{*}{\textbf{Models}} & \multirow{2}{*}{\textbf{Datasets}} & \multicolumn{3}{c|}{\textbf{Natural Language Paraphrasing}} & \multicolumn{3}{c}{\textbf{LLMTime Prediction}}  \\
    \cline{3-8}
     & & \textbf{MSE} & \textbf{MAE} & \textbf{MAPE} & \textbf{MSE} & \textbf{MAE} & \textbf{MAPE} \\ \hline
    \multirow{8}{*}{\textbf{GPT-3.5-Turbo}}
    & AirPassengers & 267.66 & 3.66 & 0.99 & 6244.07 & 61.39 & 14.43 \\
    & AusBeer & 598.45 & 5.81 & 1.36 & 841.68 & 23.59 & 5.62 \\
    & GasRateCO2 & 3.16 & 0.46 & 0.85 & 10.88 & 2.66 & 4.73 \\
    & MonthlyMilk & 968.69 & 8.61 & 1.02 & 7507.13 & 66.28 & 112.77 \\
    & Sunspots & 251.61 & 4.27 & 20.42 & 6556.55& 58.95& 217.94 \\
    (GPT-3.5-turbo-1106) 
    & HeartRate & 4.38 & 0.55 & 0.57 & 76.83 & 7.15 & 7.42 \\  
    & Istanbul-Traffic & 224.17 & 3.74 & 8.81 & 335.05 & 6.75 & 11.68\\
    & ETTh1 & 1.21 & 0.48 & 54.17 & 5.64 & 2.71 & 1.625  \\ 
    & ETTm2 & 0.81 & 0.36 & 27.33 & 3.46 & 2.17 & 1.178 \\ 
    \hline
    
    \multirow{8}{*}{\textbf{GPT-4-Turbo}} 
    & AirPassengers & 133.10 & 2.87 & 0.80 & 1286.25 & 28.04 & 6.07 \\
    & AusBeer & 661.80 & 7.24 & 1.63 & 513.49 & 18.57 & 4.28 \\
    & GasRateCO2 & 2.28 & 0.41 & 0.75 & 7.27 & 2.32 & 4.18 \\
    & MonthlyMilk & 413.63 & 4.94 & 0.57 & 4442.18 & 50.75 & 172.82 \\
    & Sunspots & 194.52 & 5.30 & 16.10 & 3374.70 & 41.87 & 321.11 \\
    (GPT-4-turbo-preview) 
    & HeartRate & 11.64 & 1.21 & 1.30 & 988.14 & 26.57 & 29.22 \\ 
    & Istanbul-Traffic & 176.91 & 3.88 & 9.67 & 195.33 & 5.53 & 10.03\\
    & ETTh1 & 1.20 & 0.49 & 47.62 & 4.73 & 1.53 & 3.282 \\ 
    & ETTm2 & 0.45 & 0.27 & 23.62 & 2.30 & 1.034 & 1.607 \\
    \hline
    
    \multirow{8}{*}{\textbf{Llama-2}} 
    & AirPassengers & 751.34 & 6.77 & 1.53 & 1317.9 & 55.49 & 11.18 \\
    & AusBeer & 591.75 & 23.25 & 5.41 & 644.82 & 17.88 & 4.08 \\
    & GasRateCO2 & 10.16 & 2.89 & 5.16 & 12.78 & 2.97 & 5.47 \\
    & MonthlyMilk & 851.17 & 84.83 & 9.46 & 3410.20 & 41.40 & 240.25 \\
    & Sunspots & 1483.29 & 33.27 & 17.79 & 4467.67 & 48.95 & 91.79 \\
    (llama-2-13B) 
    & HeartRate & 49.8 & 5.84 & 6.53 & 75.58 & 7.11 & 7.94 \\
    & Istanbul-Traffic & 306.80 & 5.39 & 7.24 & 438.28 & 7.28 & 9.81 \\
    & ETTh1 & 1.47 & 0.87 & 58.34 & 4.84 & 1.79 & 3.178 \\ 
    & ETTm2 & 0.84 & 0.41 & 29.86 & 3.31 & 2.07 & 2.153 \\ 
    \hline
    
    \multirow{8}{*}{\textbf{Gemini-Pro-1.0}} 
    & AirPassengers & 4474.54 & 31.54 & 7.02 & 6392.21 & 63.57 & 14.03 \\
    & AusBeer & 278.45 & 10.05 & 2.29 & 397.78 & 14.36 & 3.27\\
    & GasRateCO2 & 13.29 & 2.50 & 4.38 & 18.99 & 3.57 & 6.46 \\
    & MonthlyMilk & 440.29 & 11.91 & 1.39 & 628.98 & 17.01 & 1.99 \\
    & Sunspots & 438.29 & 10.47 & 1.21 & 626.03 & 14.94 & 1.73 \\
    (gemini-1.0-pro) 
    & HeartRate & 40.57 & 4.20 & 4.67 & 57.96 & 6.01 & 6.66 \\
    & Istanbul-Traffic & 267.43 & 5.69 & 8.37 & 321.56 & 7.32 & 9.71 \\
    & ETTh1 & 1.17 & 0.74 & 54.86 & 4.84 & 1.79 & 3.178 \\
    & ETTm2 & 0.88 & 0.39 & 21.82 & 3.31 & 2.07 & 2.153 \\ 
    \hline
    \end{tabular}
    \label{tab_paraseq_res}
\end{table*}

\begin{table*}[!ht]
    \vspace{-10pt}
    \caption{The results of external knowledge enhancement and baseline comparison.}
    \small
    \begin{center}
    \begin{tabular}{c|c|ccc|ccc}
    \hline
    \multirow{2}{*}{\textbf{Models}} & \multirow{2}{*}{\textbf{Dataset}} & \multicolumn{3}{c|}{\textbf{External Knowledge Enhancing}} & \multicolumn{3}{c}{\textbf{LLMTime Prediction}}  \\ \cline{3-8} 
     &  & \textbf{MSE} & \textbf{MAE} & \textbf{MAPE} & \textbf{MSE} & \textbf{MAE} & \textbf{MAPE}  \\ \hline
    & \textbf{AirPassengers} & 3713.99 & 50.37 & 10.88 & 6244.07 & 61.39 & 14.43  \\
    & \textbf{AusBeer}  & 669.01 & 21.82 & 5.12 & 841.68 & 23.59 & 5.62  \\ 
    & \textbf{GasRateCO2} & 16.47 & 3.36 & 5.97 & 10.88 & 2.66 & 4.73  \\
    & \textbf{MonthlyMilk} & 4781.26 & 55.45 & 6.25 & 7507.13 & 66.28 & 112.77  \\
    \textbf{GPT-3.5-turbo-1106} & \textbf{Sunspots} & 7072.42 & 62.61 & 194.29 & 6556.55 & 58.95 & 217.94  \\
    & \textbf{HeartRate} & 59.83 & 6.44 & 6.75 & 76.83 & 7.15 & 7.42 \\
    & \textbf{Istanbul-Traffic} & 888.31 & 28.16 & 60.11 & 1321.44 & 48.7 & 7.47 \\
    & \textbf{ETTh1} & 2.65 & 1.01 & 132.13 & 5.64 & 2.71 & 1.625  \\ 
    & \textbf{ETTm2} & 2.00 & 0.89 & 201.84 & 3.46 & 2.17 & 1.178 \\ \hline
    
    & \textbf{AirPassengers} & 1262.24 & 30.54 & 6.80 & 1286.25 & 28.04 & 6.07 \\ 
    & \textbf{AusBeer} & 345.59 & 15.70 & 3.69 & 513.49 & 18.57 & 4.28 \\ 
    & \textbf{GasRateCO2} & 6.99 & 2.29 & 4.21 & 7.27 & 2.32 & 4.18 \\ 
    & \textbf{MonthlyMilk} & 2209.33 & 44.02 & 5.12 & 4442.18 & 50.75 & 172.82 \\ 
    \textbf{GPT-4-turbo-preview} & \textbf{Sunspots} & 4571.92 & 50.24 & 334.30 & 3374.70& 41.87& 321.11\\ 
    & \textbf{HeartRate} & 78.99 & 6.96 & 7.90 & 988.14& 26.57& 29.22\\ 
    & \textbf{Istanbul-Traffic} & 954.88 & 26.92 & 47.29 & 1291.17 & 32.16 & 6.46 \\ 
    & \textbf{ETTh1} & 2.70 & 1.06 & 129.99 & 4.73 & 1.53 & 3.282 \\ 
    & \textbf{ETTm2} & 1.18 & 0.79 & 291.67 & 2.30 & 1.034 & 1.607 \\ \hline

    & \textbf{AirPassengers} & 3713.99 & 50.37 & 10.88 & 1286.25 & 28.04 & 6.07 \\ 
    & \textbf{AusBeer} & 893.56 & 21.49 & 4.87 & 644.82 & 17.88 & 4.08 \\ 
    & \textbf{GasRateCO2} & 11.38 & 3.04 & 5.49 & 12.78 & 2.97 & 5.47 \\ 
    & \textbf{MonthlyMilk} & 4722.32 & 60.36 & 7.05 & 3410.20 & 41.40 & 240.25 \\ 
    \textbf{Llama-2} & \textbf{Sunspots} & 4000.19 & 46.45 & 138.69 & 4467.67& 48.95& 91.79\\ 
    & \textbf{HeartRate} & 112.17 & 7.86 & 8.93 & 75.58& 7.11& 7.94\\ 
    & \textbf{Istanbul-Traffic} & 979.15 & 26.70 & 45.57 & 1531.37 & 34.74 & 7.42 \\ 
    & \textbf{ETTh1} & 4.15 & 1.65 & 408.11 & 4.84 & 1.79 & 3.178 \\ 
    & \textbf{ETTm2} & 3.08 & 1.47 & 810.56 & 3.31 & 2.07 & 2.153 \\ \hline

    & \textbf{AirPassengers} & 5237.85 & 51.92 & 11.08 & 6392.21 & 63.57 & 14.03 \\
    & \textbf{AusBeer} & 325.45 & 10.84 & 1.86 & 397.78 & 14.36 & 3.27 \\
    & \textbf{GasRateCO2} & 15.54 & 3.23 & 4.43 & 18.99 & 3.57 & 6.46 \\
    & \textbf{MonthlyMilk} & 491.26 & 15.18 & 1.13 & 628.98 & 17.01 & 1.99 \\
    \textbf{Gemini-1.0-pro}  & \textbf{Sunspots} & 491.64 & 11.15 & 1.27 & 626.03 & 14.94 & 1.73 \\
    & \textbf{HeartRate} & 47.45 & 4.83 & 4.67 & 57.96 & 6.01 & 6.66 \\
    & \textbf{Istanbul-Traffic} & 1253.74 & 28.25 & 5.42 & 1531.37 & 34.74 & 7.42 \\
    & \textbf{ETTh1} & 2.92 & 1.45 & 2.88 & 4.84 & 1.79 & 3.178 \\
    & \textbf{ETTm2} & 2.00 & 1.74 & 1.22 & 3.31 & 2.07 & 2.153 \\ \hline
    \end{tabular} 
    \end{center}
    \label{tab_knowledge_enhance}
\end{table*}

\section{Related Work}
\vspace{15pt}

In this section, we review two lines of research that are most relevant to ours.

\subsection{Traditional Time Series Forecasting}
\vspace{15pt}

Two commonly used traditional time series analysis methods are the ARIMA method \cite{box1970distribution} and the exponential smoothing method \cite{gardner2006exponential}. The ARIMA model is a classic forecasting method that breaks down a time series into auto-regressive (AR), difference (I), and moving average (MA) components to make predictions. On the other hand, exponential smoothing is a straightforward yet effective technique that forecasts future values by taking a weighted average of past observations. The ARIMA model requires testing data stationarity and selecting the right order. However, the exponential smoothing method is not affected by outliers; it is only suitable for stationary time series, and its accuracy in predicting future values is lower than that of the ARIMA model.

\subsection{LLMs for Time Series Forecasting}
\vspace{15pt}


The first family of methods involves either pre-training a foundational large language model or fine-tuning existing LLMs by leveraging extensive time-series data \cite{rasul2023lag, garza2023timegpt, das2023decoder,cao2023tempo}. For instance, 
\cite{rasul2023lag}
aimed to build the foundational models for time series and investigate its scaling behavior. \cite{chang2024llm4ts} proposed a two-stage fine-tuning strategy for handling multivariate time-series forecasting. Although these studies contribute significantly to understanding foundational models, they require considerable computing resources and expertise in fine-tuning procedures. Moreover, the details of the model may not be disclosed for commercial purposes \cite{garza2023timegpt,zhang2024goal}, which impedes future research. Additionally, in scenarios with limited data available, there is insufficient information for training or fine-tuning. 

In contrast, the second family of methods does not involve model parameter finetuning. 
These methods either create appropriate prompts or reprogramme inputs to effectively handle time series data 
\cite{gruver2023large, sun2023test, jin2023time, xue2023promptcast}. 
\cite{sun2023test} tokenizes the time series and manages to embed those tokens, and \cite{jin2023time} reprogrammed the time series data with text prototypes before feeding them to the LLMs. These studies illuminate the characteristics of time series data and devise methods to align them with LLMs. However, they lack an analysis of the ability and bias in forecasting time series. The most related work to us is \cite{gruver2023large}, though it lacks a quantitative analysis of the preference for the time series in LLMs, and it fails to explore the impact of input forms and prompt contents, such as converting the numerical time series into the natural language sequences and incorporating the background information into the prompt. Our work fills the gap, and we expect our work to be the benchmark for time-series analysis and provide insights for subsequent research.

\section{Conclusions and Future Work}
\vspace{15pt}

In this work, we investigate the key preferences of LLMs in the domain of time series forecasting under the zero-shot setting, revealing a proclivity for data with distinct trends and seasonal patterns. Through a blend of real and synthetic datasets, coupled with counterfactual experiments, we have demonstrated LLMs' improved forecasting performance with time series that exhibit clear periodicity. Besides, our results indicate that LLMs struggle with multi-period time series datasets, as they face difficulty in recognizing the distinct periods within them. Our findings also suggest that large language models are more sensitive to the segment of input sequences closer to the last known data than other locations. Lastly, experimental results indicate that our proposed strategies of incorporating external knowledge and transforming numerical sequences into natural language formats have yielded substantial improvements in accuracy.

\begin{table*}[!ht]
  \centering
  \caption{Summary of tests on different datasets.}
  \normalsize
  \vspace{5pt}
  \begin{tabular}{p{0.12\textwidth}|p{0.12\textwidth} p{0.12\textwidth}|p{0.08\textwidth} p{0.08\textwidth}|p{0.08\textwidth} p{0.08\textwidth}}
    \hline
    \textbf{Datasets} & \textbf{Acknowledge Test (GPT)} & \textbf{Acknowledge Test (Gemini)} & \textbf{Series Test (GPT)} & \textbf{Series Test (Gemini)} & \textbf{Dataset Detection (GPT)} & \textbf{Dataset Detection (Gemini)} \\
    \hline
    AirPassengers & Yes & Yes & Yes & No & No & No \\
    AusBeer & No & Yes & No & No & No & No \\
    GasRateCO2 & No & Yes & No & No & No & No \\
    MonthlyMilk & Yes & Yes & No & No & No & No \\
    Sunspots & Yes & Yes & No & No & No & No \\
    Wine & Yes & Yes & No & No & No & No \\
    Wooly & No & No & No & No & No & No \\
    HeartRate & Yes & Yes & No & No & No & No \\
    \hline
  \end{tabular}
  \label{table:dataset_tests}
\end{table*}

\begin{table*}[!ht]
  \centering
  \vspace{-15pt}
  \caption{Comparison of Avg Token Lengths among Original TimeLLM method, External Knowledge Enhancing and Natural Language Paraphrasing.}
  \vspace{5pt}
  \normalsize
  \begin{tabular}{l|c|c|c}
    \hline
    \textbf{Datasets} & \textbf{Avg Token Length (ori)} & \textbf{Avg Token Length (EKE)} & \textbf{Avg Token Length (NLP)} \\
    \hline
    AirPassengers & 200 & 224 & 797 \\
    AusBeer       & 200 & 220 & 797 \\
    GasRateCO2    & 200 & 211 & 797 \\
    MonthlyMilk   & 200 & 218 & 797 \\
    Sunspots      & 200 & 217 & 797 \\
    Wine          & 200 & 217 & 797 \\
    Wooly         & 200 & 216 & 797 \\
    HeartRate     & 200 & 214 & 797 \\
    \hline
  \end{tabular}
  \label{table:token_length_comparison}
\end{table*}

\begin{table*}[!ht]
  \centering
  \caption{Model performance in the analysis of LLMs' preferences.}
  \vspace{10pt}
  \normalsize
  \begin{tabular}{l|S S|S S}
    \hline
    \textbf{Dataset Name} & \textbf{GPT4-MAPE} & \textbf{GPT3.5-MAPE} & \textbf{Trend Strength} & \textbf{Seasonal Strength} \\
    \hline
    AirPassengersDataset & 6.80  & 9.98  & 1.00 & 0.98 \\
    AusBeerDataset       & 3.69  & 5.12  & 0.99 & 0.96 \\
    MonthlyMilkDataset   & 5.12  & 6.25  & 1.00 & 0.99 \\
    SunspotsDataset      & 334.30 & 194.29 & 0.81 & 0.28 \\
    WineDataset          & 10.90  & 14.98  & 0.67 & 0.92 \\
    WoolyDataset         & 20.41  & 19.26  & 0.96 & 0.82 \\
    IstanbulTrafficGPT   & 47.29  & 60.11  & 0.31 & 0.72 \\
    GasRateCO2Dataset    & 4.21   & 5.97   & 0.65 & 0.50 \\
    HeartRateDataset     & 7.90   & 6.75   & 0.42 & 0.49 \\
    TurkeyPower          & 3.36   & 3.52   & 0.90 & 0.88 \\
    \hline
  \end{tabular}
  \label{tab: Q1}
\end{table*}

\bibliographystyle{IEEEtran}
\bibliography{SigkddExp}

\end{document}